\documentclass[letterpaper, 10 pt, journal, twoside]{ieeetran}

\usepackage{cite}
\usepackage{amsmath,amssymb,amsfonts}
\usepackage{algorithmic}
\usepackage{graphicx}
\usepackage{textcomp}
\usepackage{xcolor}

\usepackage{irap_acronyms}
\usepackage{irap_acronyms2}
\usepackage{irap_math}
\usepackage{irap_SIunits}
\usepackage{irap_misc}
\makeatletter
\let\NAT@parse\undefined
\makeatother
\usepackage[numbers]{natbib}
\usepackage{subfigure}
\usepackage{multirow}
\usepackage{adjustbox}
\usepackage{lipsum}
\usepackage{mathtools}

\begin{document}

\title{OpenStreetMap-based LiDAR Global Localization in Urban Environment without a Prior LiDAR Map}

\author{Younghun Cho${}^{1}$, Giseop Kim${}^{2}$, Sangmin Lee${}^{1}$ and Jee-Hwan Ryu${}^{1*}$
\thanks{Manuscript received: September, 9, 2021; Accepted February, 7, 2022.}
\thanks{This paper was recommended for publication by Editor Sven Behnke upon evaluation of the Associate Editor and Reviewers' comments. (Corresponding author: Jee-Hwan Ryu.)
This work was supported in part by the National Research Foundation of Korea under Grant NRF-2020R1A2C200416912 and in part by the Ministry of Trade, Industry and Energy of Korea under Grant 20008548.} 

    \thanks{
      $^{1}$Younghun Cho, Sangmin Lee and Jee-Hwan Ryu are affiliated with the Department of Civil and Environmental Engineering, KAIST, Daejeon 34141, Korea {\tt\small \{lucascho, iismn, jhryu\}@kaist.ac.kr}
      \newline
      \indent $^{2}$Giseop Kim is affiliated with Autonomous Driving Group, NAVER LABS, Seongnam-si 13561, Gyeonggi-do, Korea {\tt\small giseop.kim@naverlabs.com}
    }
\thanks{Digital Object Identifier (DOI): see top of this page.}
}

\markboth{IEEE Robotics and Automation Letters. Preprint Version. Accepted February, 2022}
{Cho \MakeLowercase{\textit{et al.}}: OpenStreetMap-based LiDAR Global Localization in Urban Environment without a Prior LiDAR Map}

\maketitle

\begin{abstract}
Using publicly accessible maps, we propose a novel vehicle localization method that can be applied without using prior light detection and ranging (LiDAR) maps. Our method generates OSM descriptors by calculating the distances to buildings from a location in OpenStreetMap at a regular angle, and LiDAR descriptors by calculating the shortest distances to building points from the current location at a regular angle. Comparing the OSM descriptors and LiDAR descriptors yields a highly accurate vehicle localization result. Compared to methods that use prior LiDAR maps, our method presents two main advantages: (1) vehicle localization is not limited to only places with previously acquired LiDAR maps, and (2) our method is comparable to LiDAR map-based methods, and especially outperforms the other methods with respect to the top one candidate at KITTI dataset sequence 00.

\end{abstract}


\section{Introduction}
\label{chap:introduction}

Vehicle localization, or perceiving vehicle position, is an essential prerequisite for autonomous driving technologies. Autonomous driving algorithms, including path planning and following, cannot be accomplished without accurate vehicle position information. Owing to their high accuracy, \ac{LiDAR} sensors have been actively studied for this purpose. \ac{LiDAR}-based localization is usually performed by comparing real-time \ac{LiDAR} scans with prior \ac{LiDAR} maps created using \ac{LiDAR} scans.

However, building \ac{LiDAR} maps is a challenging: To build a \ac{LiDAR} map, a vehicle equipped with a LiDAR sensor must visit every road in the target area, scanning every road that an autonomous vehicle might be using, which is a very challenging goal. Additionally, the LiDAR maps would need to be continuously updated every time new roads are built. Furthermore, the LiDAR scans must also be appropriately collated to create the \ac{LiDAR} map. Dynamic objects and changing environments, such as passing cars and trees, make this process even more difficult, because these obstacles may hide static objects such as buildings. Thus, efforts have been made to replace LiDAR maps by finding ways to localize vehicles without visiting their real-world locations.

One such alternative approach was to use OpenStreetMap, which provides geographic, including roads and buildings, information. By utilizing static building information that does not vary in the short-term, OpenStreetMap-based descriptors can be generated at every road point to replace a \ac{LiDAR} map. Several studies have utilized OpenStreetMap for localization problems; however, most of them perform visual localization using images \cite{brubaker2013lost, panphattarasap2018automated, tangget}. Only a few studies have attempted to utilize OpenStreetMap for the vehicle localization problem, but prior attempts could not provide reliable localization results for autonomous driving \cite{yan2019global}. The research and its limitattions are reviewed in Section \ref{chap:osm_localization}.

\begin{figure}[!t]
\centering
\subfigure[Our Method]{\includegraphics[width=0.45\linewidth]{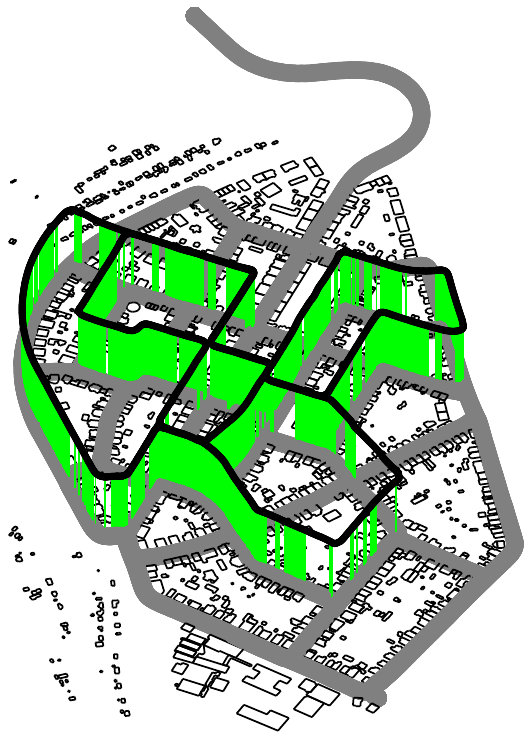}}
\subfigure[LiDAR Map-based Method]{\includegraphics[width=0.45\linewidth]{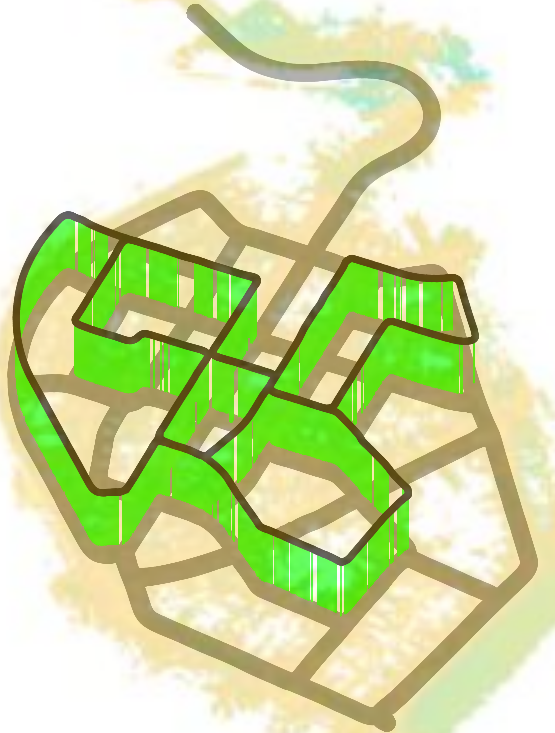}}
\caption{This paper proposes a method that performs LiDAR-based global localization without a prior LiDAR map. Based on this method, we can obtain a comparable global localization result against PointNetVLAD. In particular, our method outperforms PointNetVLAD in terms of the top one accuracy at KITTI dataset sequence 00. (a) Localization results on OpenStreetMap. (b) Localization results on the LiDAR map. Black: query LiDAR scans trajectory. Gray: reference LiDAR scans trajectory (only for (b)). Green: successful localization within 5m.}
\vspace{-0.5cm}
\label{fig:intro}
\end{figure}

This paper proposes a 1D descriptor-based vehicle localization method using OpenStreetMap without prior \ac{LiDAR} maps. Our method generates 1D descriptors, named OSM descriptors, from both OpenStreetMap information and \ac{LiDAR} scans acquired from vehicles, to determine the vehicle's position by comparing those descriptors. In this process, a set of descriptors generated from OpenStreetMap replaces the conventional LiDAR maps. \figref{fig:intro} shows that the OSM descriptor performs comparably with PointNetVLAD on the KITTI dataset sequence 00. Moreover, the OSM descriptor is complicated enough to discriminate between places without the adapting additional frameworks, such as Monte Carlo Localization.

The contribution of this paper is as follows:
\begin{itemize}
\item To the best of our knowledge, our method is the first attempt for LiDAR kidnapped place recognition using OpenStreetMap and shows localization performance comparable to other methods using prior LiDAR maps.
\item As opposed to conventional methods, our method is not limited to only places that are covered by LiDAR maps.
\item Unlike conventional OpenStreetMap-based LiDAR localization methods, our method does not require additional processes such as a particle filter.
\end{itemize}

\section{Related Works}
\label{chap:related_works}

\subsection{Descriptor-based LiDAR Localization}
LiDAR is the most widely used sensor for vehicle localization research. There are two forks of research that utilized LiDAR sensor: registration-based methods and feature-based methods. Registration-based methods, or dense methods, utilize all points in the point clouds for matching. In contrast, feature-based methods, or sparse methods, extract features from point clouds and match them with prior LiDAR maps. Methods using descriptors can also be considered as feature-based methods.

\citeauthor{himstedt2014large}, \citeauthor{he2016m2dp}, and \citeauthor{kim20191} developed 2D descriptors called GLARE, M2DP, and Scan context respectively to convert point clouds into descriptors \cite{himstedt2014large, he2016m2dp, kim2018scan, kim2021scan, kim20191}. GLARE transforms 2D LiDAR scans into histograms and uses the bag-of-words framework \cite{himstedt2014large}. M2DP projects 3D point clouds into multiple 2D planes and uses their densities as descriptors, which are more robust even with the presence of noisy point clouds \cite{he2016m2dp}. The Scan context extracts the highest point for every angle and calculates the distance from the point clouds, outperforming other descriptor-based methods \cite{kim2018scan, kim2021scan}.

Recently, deep learning-based methods have been widely applied to LiDAR localization. The most representative works include L3-Net and PointNetVLAD \cite{lu2019l3, uy2018pointnetvlad}. L3-Net was the first learning-based LiDAR localization method and was developed into the DeepVCP \cite{lu2019deepvcp}. As can be inferred from the title of the paper, PointNetVLAD is a combination of two methods: PointNet and NetVLAD \cite{qi2017pointnet, arandjelovic2016netvlad}. This method achieves permutation invariance, because NetVLAD is a symmetric function. Additionally, the researchers adopted a lazy triplet loss function that maximizes the difference between descriptors.

However, these LiDAR localization methods rely on prior LiDAR maps, because they compare point clouds or features with prior LiDAR maps, which are very limited.

\subsection{OpenStreetMap-based Localization}
\label{chap:osm_localization}
Therefore, several researchers sought to localize autonomous vehicles using OpenStreetMap. Most OpenStreetMap-based localization methods utilized visual sensors rather than LiDARs, or overhead images rather than OSM data, for their works \cite{brubaker2013lost, panphattarasap2018automated, tangget}. Furthermore, as mentioned earlier, there have been a few attempts to achieve LiDAR localization using OpenStreetMap. \citeauthor{floros2013openstreetslam} fitted trajectories generated from visual odometry to OpenStreetMap road information using Chamfer matching \cite{floros2013openstreetslam}. \citeauthor{ruchti2015localization} also fitted the calculated trajectory to OpenStreetMap road information, but it used the LiDAR point clouds instead of visual sensors. Road and non-road road cells from LiDAR point clouds were classified, and the output was used as the weights of the Monte Carlo Localization \cite{ruchti2015localization}. However, these methods will not function accurately when a vehicle is driven straight for a long time, because they only use the topological information of road networks.

In contrast, \citeauthor{vysotska2016exploiting} utilized the building information instead of the road information to localize vehicles \cite{vysotska2016exploiting}. LiDAR point clouds were fitted to the OpenStreetMap building information, and the results were used as constraints for graph-based SLAM. \citeauthor{suger2017global} classified the semantic terrain information of the surrounding environment using LiDAR point clouds, and used the result for the Monte Carlo Localization's weighting function \cite{suger2017global}. \citeauthor{yan2019global} suggested 4-bit semantic descriptors, classifying the existence of buildings and roads in four directions, and performing Monte Carlo Localization using their descriptors \cite{yan2019global}. However, the results of this method are not sufficiently accurate for autonomous driving, because the resulting trajectories pass through buildings.

Another major constraint of these localization methods is that they cannot be used independently, but must be combined with other methods such as Monte Carlo Localization or graph-based SLAM. Additionally, because of the fundamental problem of a Monte Carlo Localization, time is required to initialize the vehicle's position. On the contrary, our method is a stand-alone method for kidnapped localization problems that can be used independently from other methods.

\section{Vehicle Localization using OSM Context}
\label{chap:methods}

\begin{figure*}[!t]
\centerline{\includegraphics[width=0.95\textwidth]{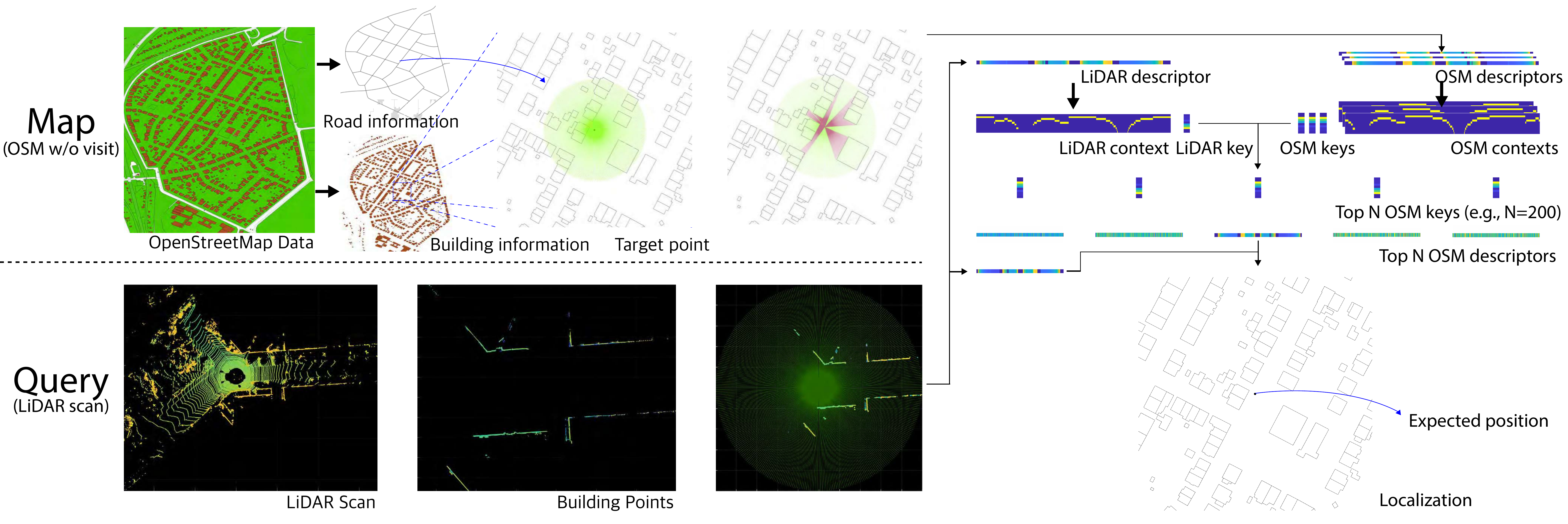}}
\vspace{-0.4cm}
\caption{Overall pipeline of the proposed method. We generated a reference map using the OpenStreetMap data by converting the building information into OSM descriptors and OSM keys. For every query LiDAR scan, we converted it into a LiDAR descriptor and LiDAR key. Comparing them in a 2-stage method, we can find the most similar OSM descriptor to localize the vehicle's position.}
\vspace{-0.5cm}
\label{fig:pipeline}
\end{figure*}

The pipeline of the proposed method is visualized in \figref{fig:pipeline}. This chapter states the problem definition and describes the methods used in this paper, which consist of four modules: OSM descriptor generation, LiDAR descriptor generation, rotation invariant key generation, and finding the most similar descriptor.

\subsection{Problem Definition}
This paper focuses on the problem of LiDAR localization without prior visits to the target area. Specifically, we build a reference map consisting of OSM descriptors at each position in the target area, without visiting the actual location. Because LiDAR scans contain comprehensive information about the surrounding environment, while OSM contains only building information, only the building points should be extracted from LiDAR scans to warrant an equal comparison. We converted each extracted building point from the LiDAR scans into 1D descriptors, which have the same dimensions as the OSM descriptor, and found the descriptor most similar to the reference map. We assume that the ground is locally flat and buildings are not necessarily rectangular but structured and have straight vertical walls.

Denoting the function that converts OpenStreetMap building information, $b_n$, into OSM descriptor as $f(.)$, the reference map is built as a set of OSM descriptors $M = \{m_1, m_2, ..., m_N\}$ where $m_n = f(b_n)$. Given a query point cloud $q$, our aim is to find an OSM descriptor, $m_\ast$, with the minimum distance.

To solve this problem, we extracted a building point cloud, $q^\prime = R(q)$, where $R(.)$ is a sematic segmentation network using RangeNet++ \cite{milioto2019iros}. Using the function $f(.)$, we generated a LiDAR descriptor $\bar{q} = f(q^\prime)$. By comparing $\bar{q}$ and $M$, we found the OSM descriptor with the minimum distance using the distance function $d(.)$, where $d(\bar{q}, m_\ast) \leq d(\bar{q}, m_n), n \leq N$. In this paper, we used the L1-norm as the distance function $d(.)$.

\subsection{OSM Descriptor Generation}
\label{chap:OSM_descriptor_generation}

\begin{figure}[!t]
\centering
\subfigure[]{\includegraphics[width=0.3\linewidth, height=0.32\linewidth]{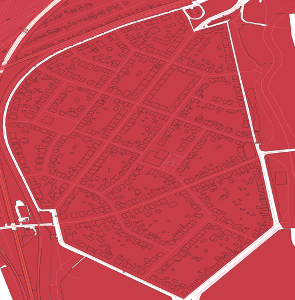}}
\subfigure[]{\includegraphics[width=0.3\linewidth, height=0.32\linewidth]{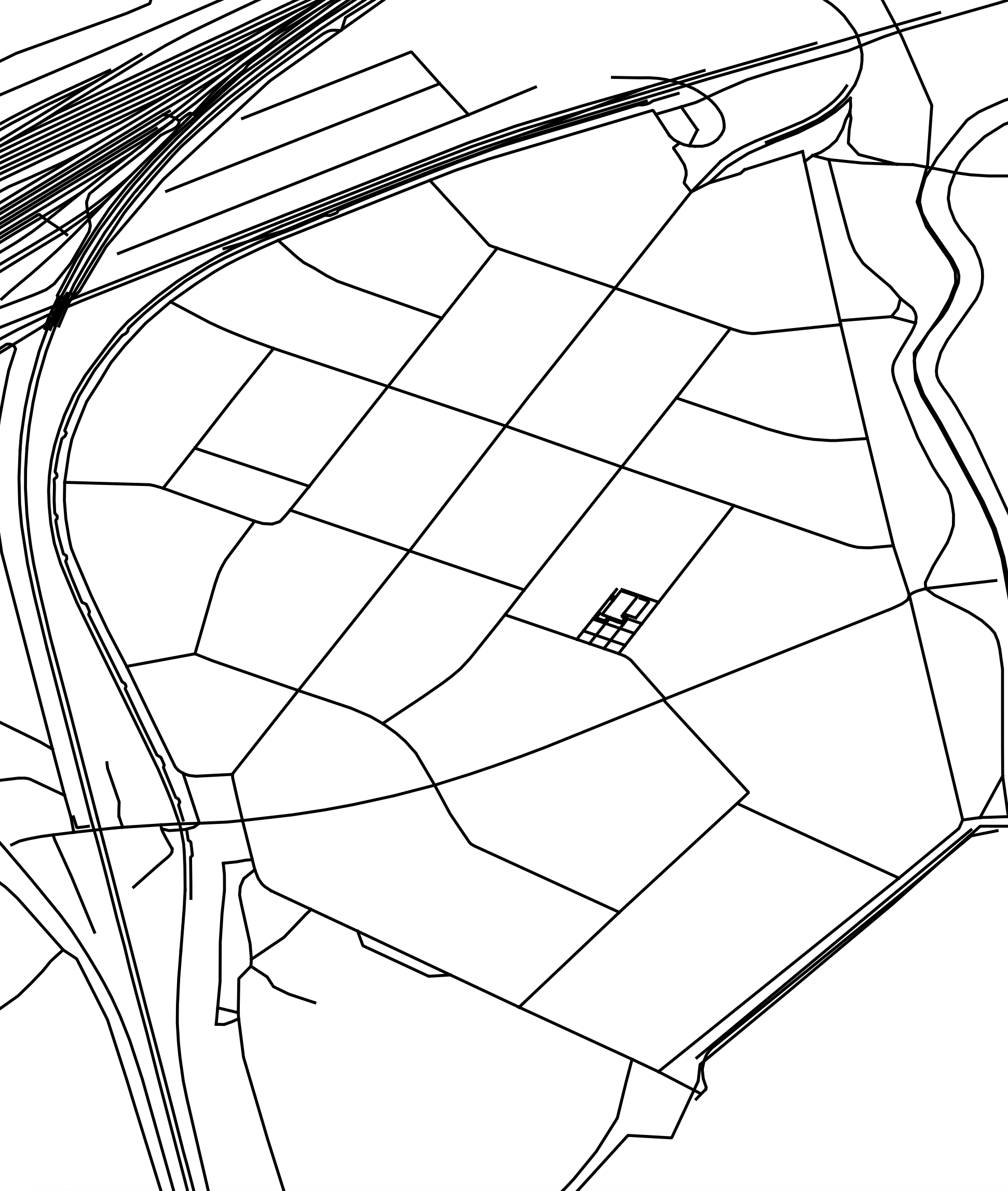}}
\subfigure[]{\includegraphics[width=0.3\linewidth, height=0.32\linewidth]{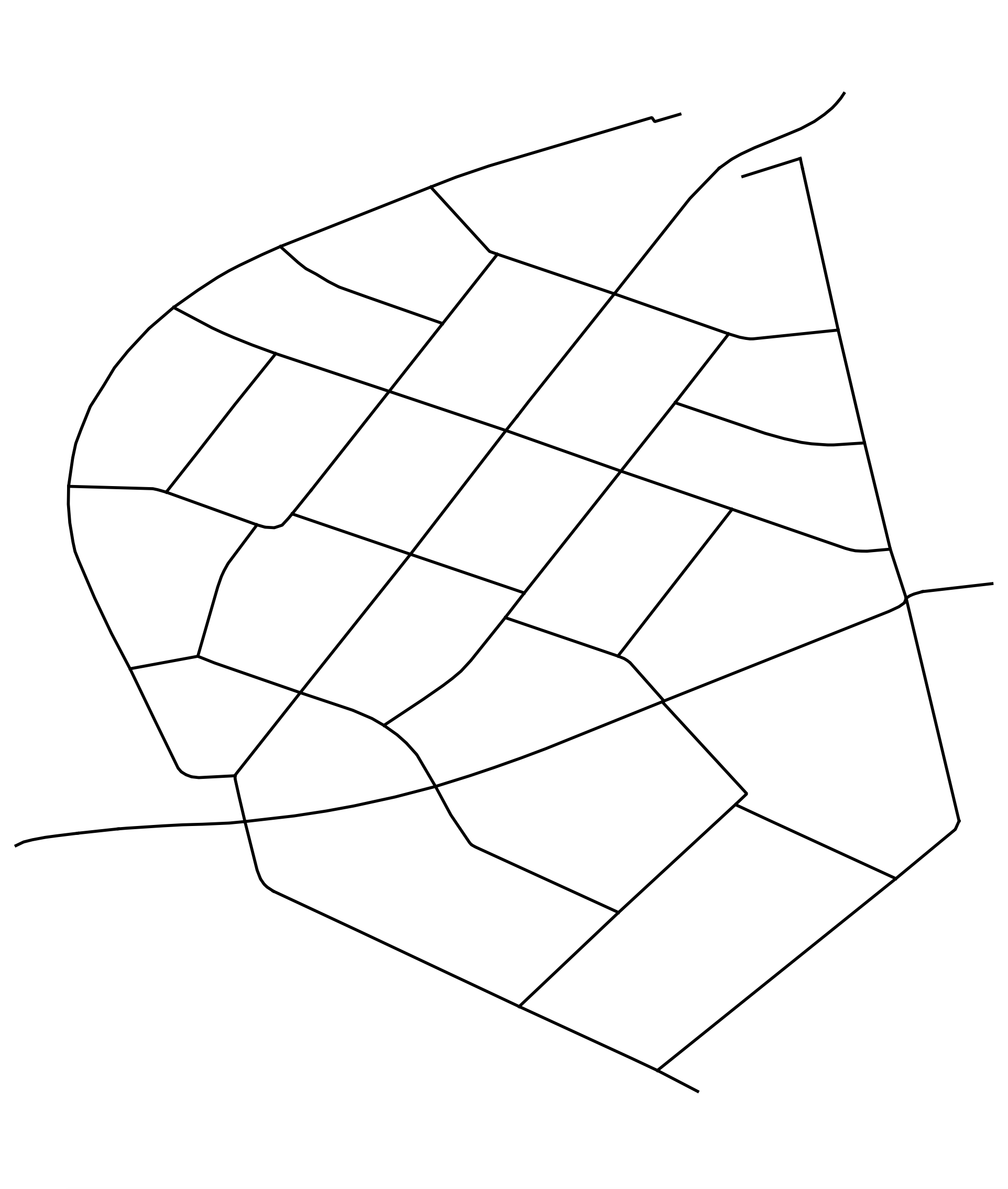}}
\caption{OSM data visualization. (a) Entire information provided by OSM. (b) ‘lines’ layer of OSM. (c) Extracted road information and interpolated (x,y) points.}
\label{fig:OSM_vis}
\end{figure}

The OSM descriptor is a 1D descriptor that is designed to replace \ac{LiDAR} maps in the vehicle localization scheme. The OSM descriptor mimics a 2D LiDAR scan using the OpenStreetMap building information, and the set of OSM descriptors in the target area can replace the \ac{LiDAR} map.

An OSM descriptor is generated by calculating the distance to the buildings using OpenStreetMap data. OpenStreetMap provides five layers of information: ‘points’, ‘lines’, ‘multiline strings, ‘multi polygons’, and ‘other relations’. ‘multi polygons’ layer provides building information as tuples that represent building edges, and ‘lines’ layer provides roads, rivers, and other line information. Among the ‘lines’ layer, road information is labeled as ‘highway’ and provides sparse points that are just enough to express the road shape as shown in \figref{fig:OSM_vis}. Therefore, we interpolated road points with a 1-meter interval and generated an OSM descriptor for every interpolated point $(x, y)$ in UTM converted coordinate. Denoting a tuple of the building edges $t = (X_1, X_2) = ((x_1, y_1), (x_2, y_2))$ and the target position in the UTM coordinate as $(x, y)$, we find the shortest distance to a building $d_n^\ast$ for each search angle $\theta_n$ using \eqref{eq:dist_to_building}.

\begin{figure}[!t]
\centering
\subfigure[]{\includegraphics[width=0.4\linewidth]{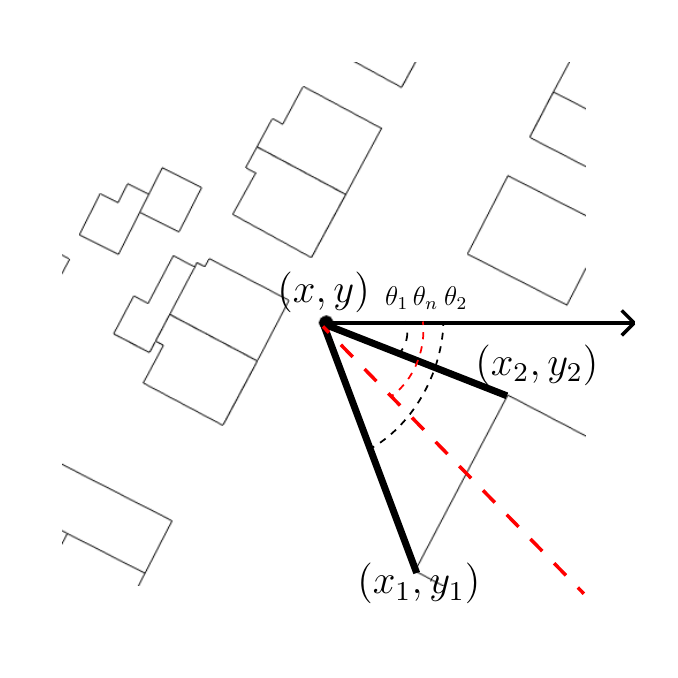}}
\subfigure[]{\includegraphics[width=0.4\linewidth]{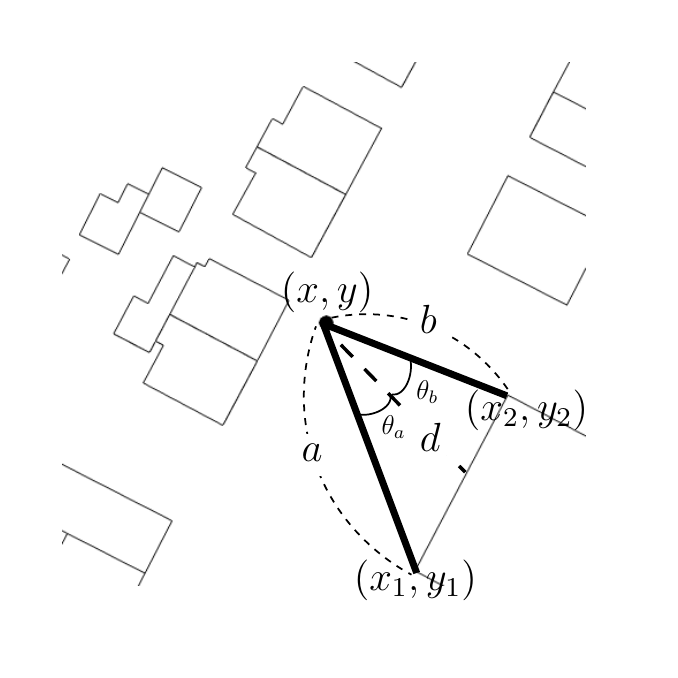}}
\caption{(a) Distance is calculated only if the target angle is in between two vertices of the building edge. (b) Distance to building is calculated using the fact that a sum of two small triangle areas is equal to the area of a large triangle.}
\label{fig:distance}
\end{figure}

\begin{equation}
d_n^\ast \leq d_n = \frac{ab\sin(\theta_a+\theta_b)}{a\sin(\theta_a)+b\sin(\theta_b)}, \quad \theta_1 \leq \theta_n \leq \theta_2,
\label{eq:dist_to_building}
\end{equation}
where $a$ and $b$ denote the distance between the current position and $t$. $\theta_1$ and $\theta_2$, respectively, denote the smaller and larger angle between the horizontal axis and the line segments joining the target point $(x, y)$ and building vertices $(x_1, y_1)$ and $(x_2, y_2)$, and $\theta_a = \theta_n - \theta_1$ and $\theta_b = \theta_2 - \theta_n$ denote the angles between the target angle and the angle to $t$, as shown in \figref{fig:distance}. If there are no buildings at the target angle, $d_\ast$ is considered as zero. $d_n$ is calculated when $\theta_1 \leq \theta_n \leq \theta_2$, which means that the target angle is in between the two vertices of the building edge. By stacking the calculated $d_n^{\ast}$ from all angles, the OSM descriptor was defined as $m_n = \{d_1^\ast, d_2^\ast, ..., d_{360}^\ast\}$.

\subsection{LiDAR Descriptor Generation}
\label{chap:LiDAR_descriptor_generation}

\begin{figure}[!t]
\centering
\subfigure[Original point cloud]{\includegraphics[width=0.46\linewidth]{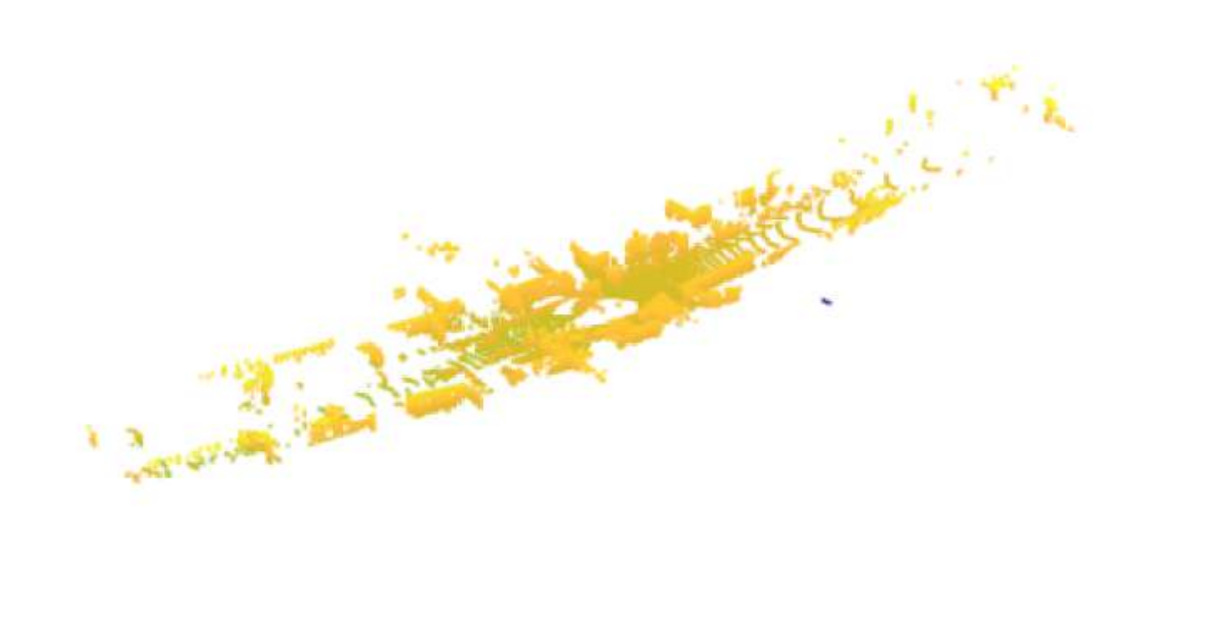}}
\subfigure[Building point cloud]{\includegraphics[width=0.46\linewidth]{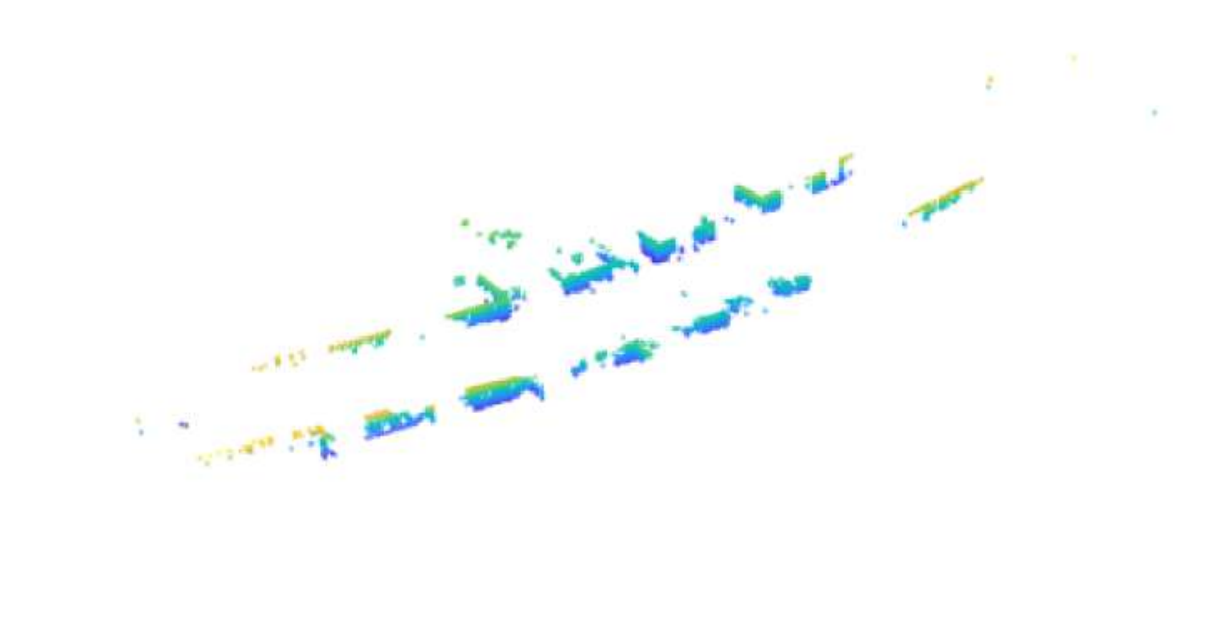}} \\
\subfigure[LiDAR descriptor]{\includegraphics[width=0.95\linewidth]{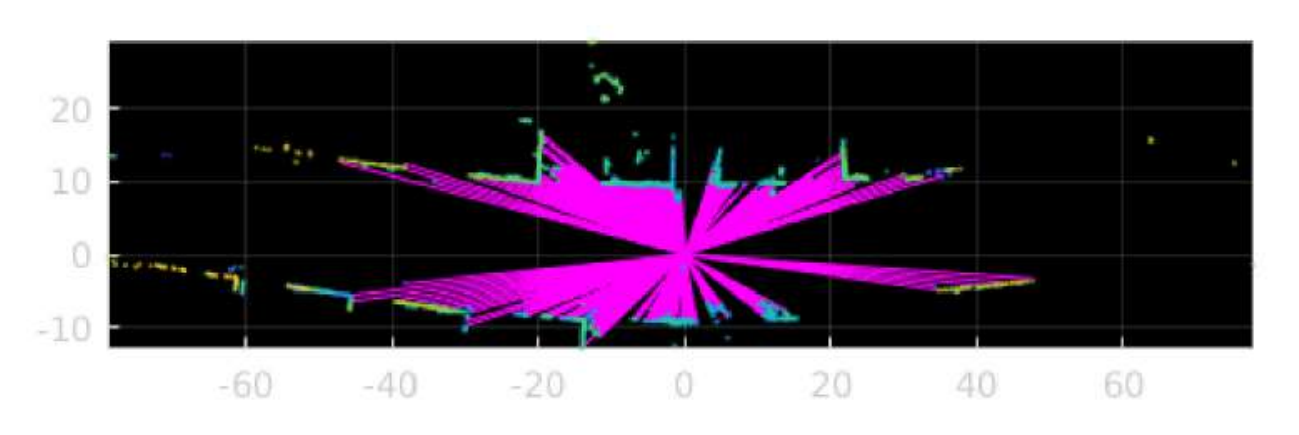}}
\caption{Visualization of LiDAR descriptor generation process.}
\label{fig:pc_descriptor}
\end{figure}

To compare the \ac{LiDAR} scans with the OSM descriptors, \ac{LiDAR} scans should be converted into the same format as the OSM descriptors. First, we extracted the building points in a \ac{LiDAR} scan using RangeNet++ \cite{milioto2019iros}, because the OSM descriptors are made using building information only. However, \ac{LiDAR} scans consist of points, whereas OpenStreetMap building information consists of the vertices of the buildings only. Therefore, a query \ac{LiDAR} descriptor $\bar{q} = \{d_1^\ast,d_2^\ast, ..., d_N^\ast\}$ can be generated by calculating the shortest distance $d_n^\ast \leq d_n$ to building points $P = \{X_1, X_2, ..., X_N\}$ at each search angle, where $d_n$ denotes the distance to the point $X_n$. The entire process for Section \ref{chap:LiDAR_descriptor_generation} is shown in \figref{fig:pc_descriptor}.

\subsection{Rotation-Invariant Descriptor Generation}

Because vehicles do not always travel in one direction only, there can be multiple scenarios of cars in a spot: cars can come in from two opposite roads, or even from multiple roads, in the case of an intersection. To account for localization in all of these cases, the building rotation-invariant descriptors are pivotal. Here, we introduce a method for making a previously described descriptor invariant to rotation.

\begin{figure}[!t]
\centerline{\includegraphics[width=0.95\linewidth]{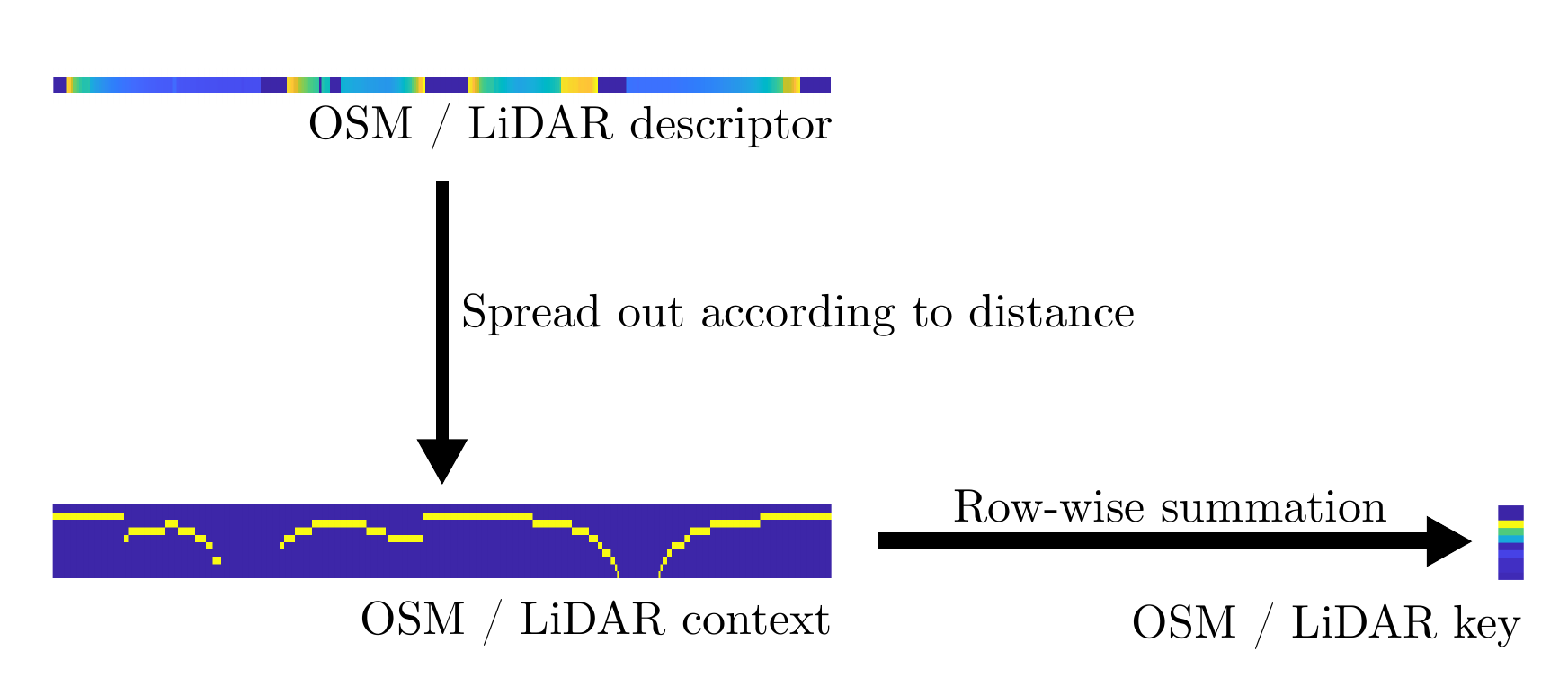}}
\caption{rotation invariant key generation}
\label{fig:rot_inv_descriptor}
\end{figure}

This can be accomplished simply by summing or averaging the features of all the angles. However, adding all 1-dimensional vectors yield only one value, which can diminish the discriminatory power as a descriptor. Therefore, to avoid the loss, we transform the OSM descriptors and LiDAR descriptors that are 1-dimensional vectors into 2-dimensional context $\hat{q}^\prime$ by setting $\lceil \bar{q}(i)/l_b\rceil$-th element of every integer angle $i$-th column to be one and others to be zero where $l_b$ denotes the length of each bin of the descriptor as shown in \eqref{eq:1d_to_2d}. In this paper, we used $l_b = 5\,m$ because 5$\,$m bin performed better compared to 2$\,$m and 10$\,$m bin as shown in \tabref{tab:KITTI_result}.

\begin{equation}
\hat{q}^\prime(k,i) = \begin{cases}
&1, \quad k = \lceil\bar{q}(i)/l_b\rceil \\
&0, \quad others
\end{cases}
, \quad 1 \leq i \leq 360, \quad i \in Z,
\label{eq:1d_to_2d}
\end{equation}
where $\bar{q}(i)$ denotes $i$-th value of the LiDAR descriptor $\bar{q}$, and $\lceil * \rceil$ is a ceil operator. By summing up the features of this 2-dimensional context for all angles at each row (i.e., interval) using \eqref{eq:2d_to_rotinv}, the rotation-invariant descriptor OSM keys and LiDAR keys are built as shown in \figref{fig:rot_inv_descriptor}.

\begin{equation}
\hat{q}(d) = \sum_{i=1}^{360}\hat{q}^\prime(d,i), \quad 1 \leq d \leq R/l_b, \quad d \in Z,
\label{eq:2d_to_rotinv}
\end{equation}
where $R$ denotes the maximum sensor range. In this paper, we used $R=50$. For OSM descriptors, we can apply the same process to make rotation-invariant OSM keys $\hat{M} = \{\hat{m_1}, \hat{m_2}, ..., \hat{m_N}\}$.

\begin{figure}[!t]
\centerline{\includegraphics[width=0.95\linewidth]{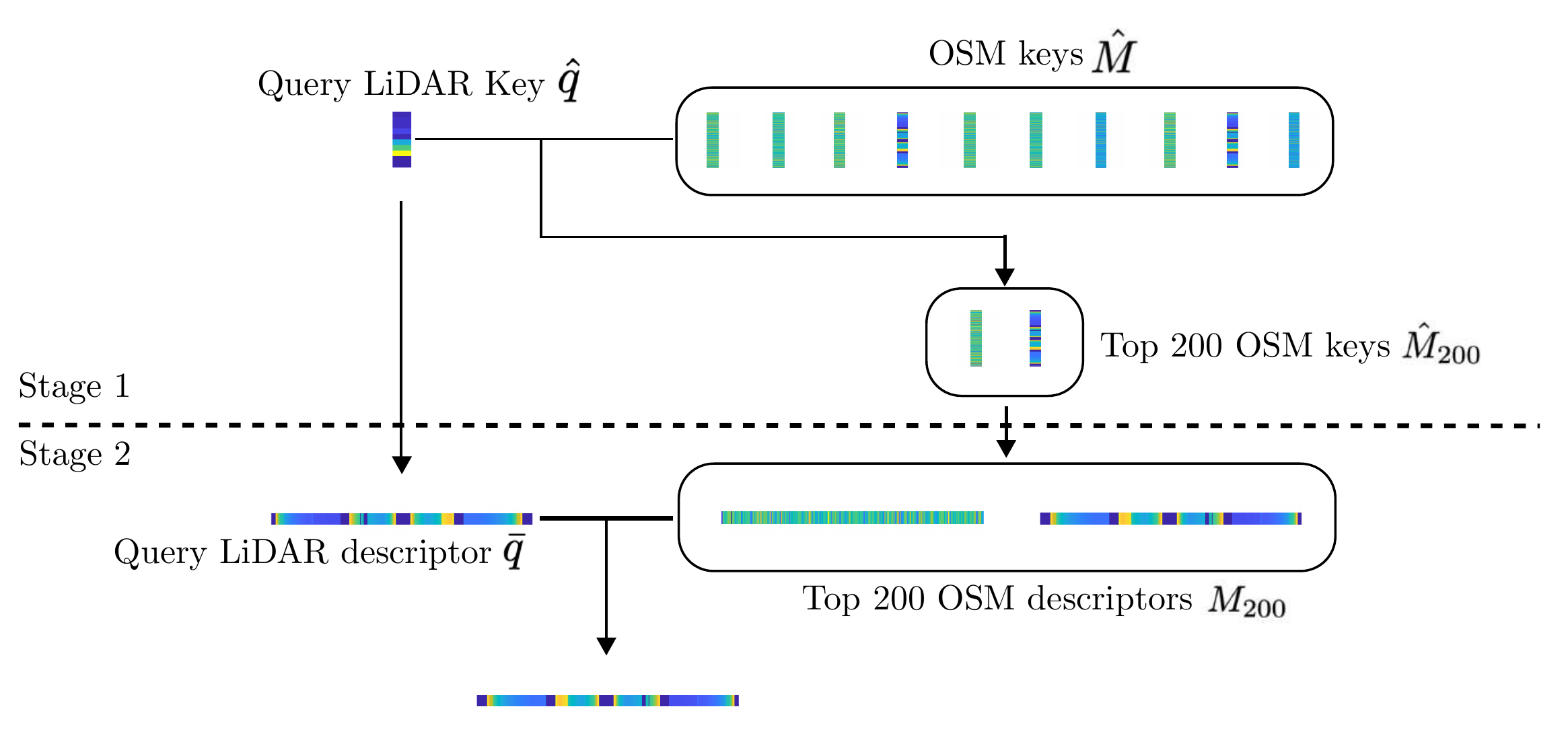}}
\caption{Pipeline of our 2-stage method. In the first stage, we sort OSM keys that have smaller dimensions compared to the OSM descriptors for performing faster calculations. We selected 200 keys; we could only compare 200 descriptors among 14000 candidates in the KITTI dataset sequence 00.}
\label{fig:2_stage_method}
\end{figure}

\begin{table*}[!thbp]
\caption{Information of Datasets used in this Paper}
\begin{center}
\begin{tabular}{c||ccccccccc}
\hline
\textbf{Dataset} & & \multicolumn{8}{c}{\textbf{Sequence}} \\
\hline
\multirow{3}{*}{KITTI} & Name & \texttt{00} & \texttt{02} & \texttt{05} & \texttt{06} & \texttt{07} & \texttt{08} & \texttt{09} & \texttt{10} \\
\cline{3-10}
& $\#$ LiDAR scan & 4541 & 4661 & 2761 & 1101 & 1101 & 4071 & 1591 & 1201 \\
& $\#$ OSM contexts & 7010 & 19523 & 11074 & 11074 & 7010 & 8250 & 11036 & 11036 \\
\hline
\multirow{3}{*}{KITTI-360} & Name & \multicolumn{2}{c}{\texttt{00}} & \multicolumn{2}{c}{\texttt{05}} & \multicolumn{2}{c}{\texttt{06}} & \multicolumn{2}{c}{\texttt{09}} \\
\cline{3-10}
& $\#$ LiDAR scan & \multicolumn{2}{c}{11518} & \multicolumn{2}{c}{6743} & \multicolumn{2}{c}{9699} & \multicolumn{2}{c}{14056} \\
& $\#$ OSM contexts & \multicolumn{2}{c}{10528} & \multicolumn{2}{c}{13512} & \multicolumn{2}{c}{9363} & \multicolumn{2}{c}{7664} \\
\hline
\end{tabular}
\label{tab:dataset}
\end{center}
\end{table*}

\subsection{Finding the Most Similar Descriptor}

Finally, we were able to localize a vehicle by comparing the query LiDAR descriptor $\bar{q}$ and the set of OSM decriptors $M$ in two stages, as shown in \figref{fig:2_stage_method}. In the first stage, we compared the query LiDAR key $\hat{q}$ of the query LiDAR descriptor $\bar{q}$ with the set of OSM keys $\hat{M}$ to rank and extract the top 200 OSM keys $\hat{M}_{200}$ with the highest similarity using the L1-norm. The 1-stage method provides sorted candidates for the second stage.

Subsequently, in the second stage, the top 200 OSM descriptors $M_{200}$ corresponding to the $\hat{M}_{200}$ are compared once again to the query LiDAR descriptor $\bar{q}$, to re-rank the descriptors. At the second stage, $\bar{q}$ and $M$ are no longer rotation invariant. However, we compared them by rotating $\bar{q}$ for every angle. Although the comparison itself is a time-consuming process, it can be done within $0.0624s$ per frame because we have done this only with the top 200 candidates. This two-step method exhibits not only a high speed, achieved with initial screening using rotation-invariant descriptors, but also a high accuracy, achieved in comparison with the OSM descriptors. If the location of the highest-ranked OSM descriptors and that of the query LiDAR descriptor is within 5$\,$m, the result is considered as a success.

\section{Experiments}
\label{chap:experiments}

\subsection{Benchmark Datasets}
We used two of the publicly available datasets that included LiDAR scans as query LiDAR scans: KITTI \cite{Geiger2013IJRR} and KITTI-360 \cite{Xie2016CVPR}. Both provide several sequences containing LiDAR scans, images, and poses of residential areas. In this paper, we utilized LiDAR scans and poses as query data. We converted all of the LiDAR scans into LiDAR descriptors using the method illustrated in Section \ref{chap:LiDAR_descriptor_generation}. Information of the KITTI and KITTI-360 datasets are summarized in \tabref{tab:dataset}.

For the reference map, we used OpenStreetMap, which provides ‘points’, ‘lines’, ‘multiline strings, ‘multi polygons’, and ‘other relations’ data. Among these data, we utilized lines and multipolygons that represent lines, such as roads and rivers, and polygons, such as buildings and districts on the map. By filtering them, we extracted the road and building information. Because road data are inconsistent and sparse, interpolation was necessary to make them consistent and dense. We used the 1$\,$m interpolation for this experiment. For each KITTI and KITTI-360 dataset sequence, we exported OpenStreetMap data of a rectangular area that encompass the entire trajectory of the sequence. Using exported data, we generated OSM descriptors for each location with interpolated road and building data using the method illustrated in Section \ref{chap:OSM_descriptor_generation}.

\begin{table*}[!t]
\caption{KITTI and KITTI-360 Localization Accuracy (\%)}
\begin{center}
\begin{adjustbox}{width=0.9\linewidth}
\begin{tabular}{llcccccccccccc}
\hline
\multicolumn{2}{c}{\multirow{3}{*}{Method ($l_b$)}} & \multicolumn{8}{c}{\texttt{KITTI}} & \multicolumn{4}{c}{\texttt{KITTI-360}}\\
\cline{3-14}
& & \multirow{2}{*}{\texttt{00}} & \multirow{2}{*}{\texttt{02}} & \multirow{2}{*}{\texttt{05}} & \multirow{2}{*}{\texttt{06}} & \multirow{2}{*}{\texttt{07}} & \multirow{2}{*}{\texttt{08}} & \multirow{2}{*}{\texttt{09}} & \multirow{2}{*}{\texttt{10}} & \multirow{2}{*}{\texttt{00}} & \multirow{2}{*}{\texttt{05}} & \multirow{2}{*}{\texttt{06}} & \multirow{2}{*}{\texttt{09}} \\
&&&&&&&&&&&&&\\
\hline                                     
\multirow{3}{*}{\textbf{Ours (2m)}} & top1  & 45.72          & \textbf{2.45} & 36.87          & 59.67          & 37.06          & 27.09          & 19.80          & 11.49          & \textbf{40.04} &  8.36          &  \textbf{8.97} & 34.40          \\
                                    & top5  & 54.22          & \textbf{3.15} & \textbf{45.31} & 65.40          & \textbf{51.04} & 35.37          & 24.07          & 12.66          & \textbf{47.78} & 11.51          & \textbf{11.56} & 42.72          \\
                                    & top10 & 57.74          & \textbf{3.43} & \textbf{50.49} & 68.12          & 54.22          & 38.79          & 25.83          & \textbf{13.66} & \textbf{52.34} & 13.76          & \textbf{13.45} & 46.74          \\
\hline
\multirow{3}{*}{\textbf{Ours (5m)}} & top1  & \textbf{48.34} & 1.85          & \textbf{37.52} & \textbf{62.13} & \textbf{37.33} & \textbf{29.89} & \textbf{22.25} & \textbf{11.66} & 39.89          &  \textbf{9.31} &  8.35          & \textbf{35.64} \\
                                    & top5  & \textbf{56.86} & 2.36          & 44.62          & \textbf{67.03} & 50.77          & \textbf{36.70} & \textbf{26.21} & \textbf{12.99} & 46.67          & \textbf{12.62} & 10.81          & \textbf{44.61} \\
                                    & top10 & \textbf{61.15} & 2.64          & 49.51          & \textbf{70.75} & \textbf{56.49} & \textbf{39.50} & \textbf{29.23} & 13.57          & 50.55          & \textbf{15.10} & 12.76          & \textbf{48.78} \\
\hline
\multirow{3}{*}{\textbf{Ours (10m)}} & top1  & 44.13         & 1.37          & 30.32          & 57.77          & 36.60          & 25.84          & 21.68          & 11.07          & 37.70          &  7.50          & 7.45           & 32.42          \\
                                     & top5  & 52.87         & 1.72          & 35.97          & 61.58          & 50.40          & 32.60          & 26.15          & 12.32          & 43.79          & 10.08          & 9.89           & 40.59          \\
                                     & top10 & 56.20         & 1.93          & 38.90          & 64.49          & 54.59          & 35.64          & 28.41          & 12.99          & 47.24          & 12.03          & 11.58          & 44.62          \\
\hline
\end{tabular}
\end{adjustbox}
\label{tab:KITTI_result}
\end{center}
\end{table*}

\vspace{-0.2cm}

\begin{figure*}[!t]
\centerline{\includegraphics[width=0.95\textwidth]{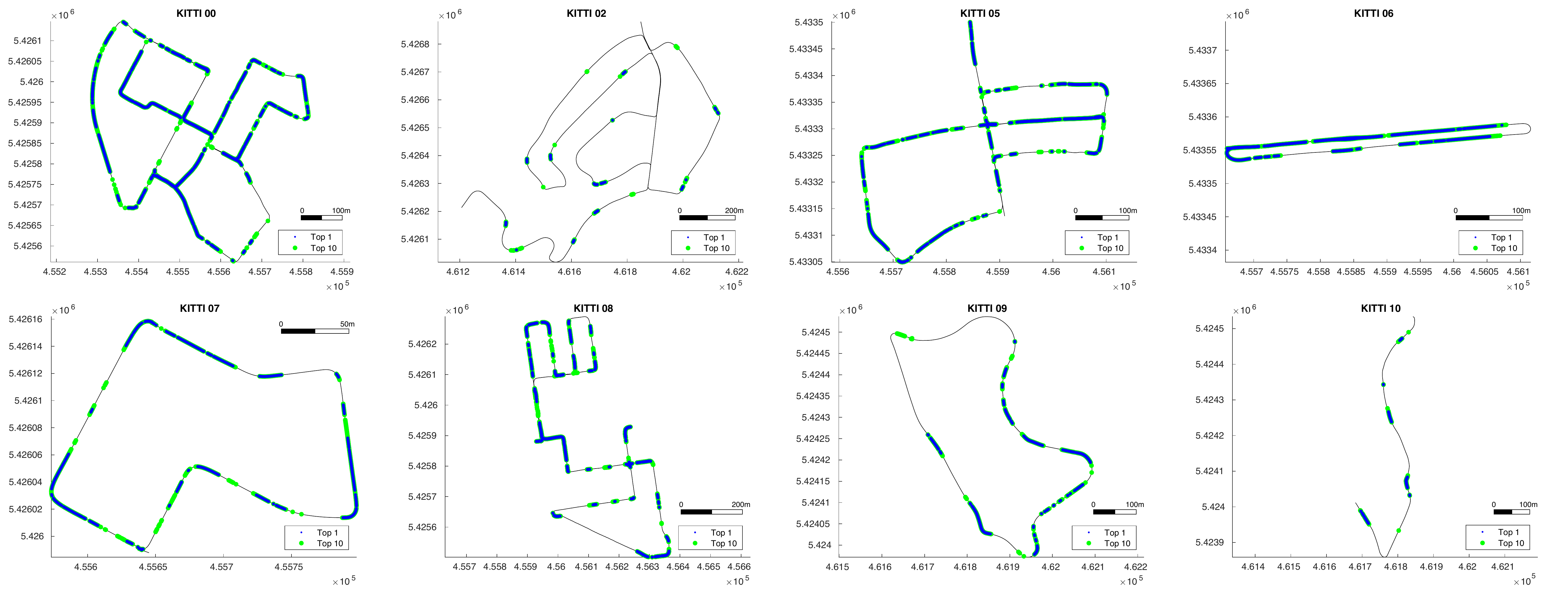}}
\caption{Top 10 accuracies of the KITTI dataset sequences. Blue: successful localization with the top one candidate. Green: successful localization with the top 10 candidates.}
\label{fig:kitti_results}
\end{figure*}

\vspace{-0.2cm}
\begin{figure*}[!t]
\centerline{\includegraphics[width=0.95\textwidth]{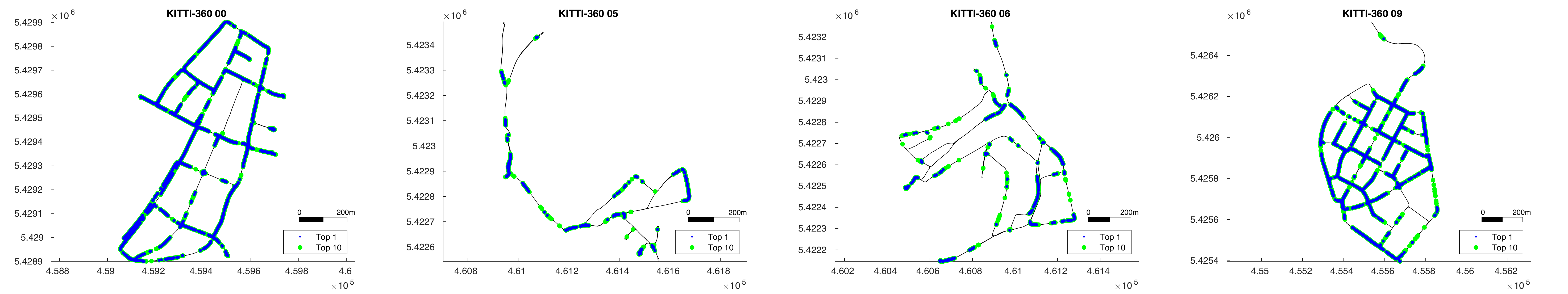}}
\caption{Top 10 accuracies of the KITTI-360 dataset sequences. Blue: successful localization with the top one candidate. Green: successful localization with the top 10 candidates.}
\label{fig:kitti360_results}
\end{figure*}

\vspace{-0.2cm}

\subsection{Comparison Methods}
\label{chap:comparison_methods}

We compared our method with PointNetVLAD \cite{uy2018pointnetvlad}, a state-of-the-art LiDAR point cloud retrieval method. We used the same parameter setting as PointNetVLAD: use ground-removed point cloud, point cloud within 25$\,$m, normalized it into with zero as the mean and range between $[-1, 1]$, and 4096 points. Using the pre-trained model provided by the author, we converted LiDAR scans into 256-dimensional descriptors.

We only compared the performances at the KITTI dataset sequence 00, because while our method can perform a LiDAR localization at every sequence, PointNetVLAD can be performed only if one sequence fully encompasses the other sequence. In the KITTI and KITTI-360 datasets, only KITTI-360 sequence 09 encompasses KITTI sequence 00. Therefore, we compared the results of PointNetVLAD using KITTI sequence 00 as query LiDAR scans and KITTI-360 sequence 09 as the reference map.

\section{Results}
\label{chap:results}

In this section, we present the results of analyzing the performance of our method in an outdoor large-scale LiDAR localization, especially in kidnapped situations. We compared the performance of PointNetVLAD, which also performed outdoor large-scale LiDAR localizations. We evaluated the accuracies of each method using the top one, top five, and top 10 candidates, to show that our method is robust even with a smaller number of candidates. In particular, we describe how our method is specialized for autonomous driving situations, because it has a rotation-invariant characteristic that can distinguish intersections and shows higher accuracy with a smaller number of candidates.

\subsection{LiDAR Localization Performance on the KITTI dataset}
\begin{figure}[!t]
\centering
\subfigure[Query point cloud]{\includegraphics[width=0.46\linewidth]{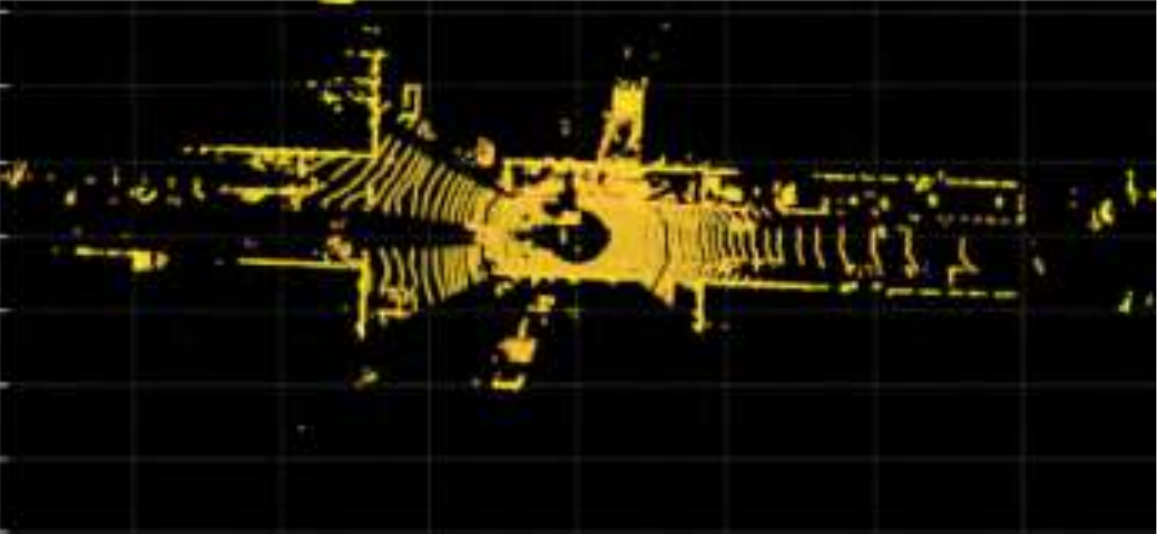}}
\subfigure[Query LiDAR descriptor]{\includegraphics[width=0.46\linewidth]{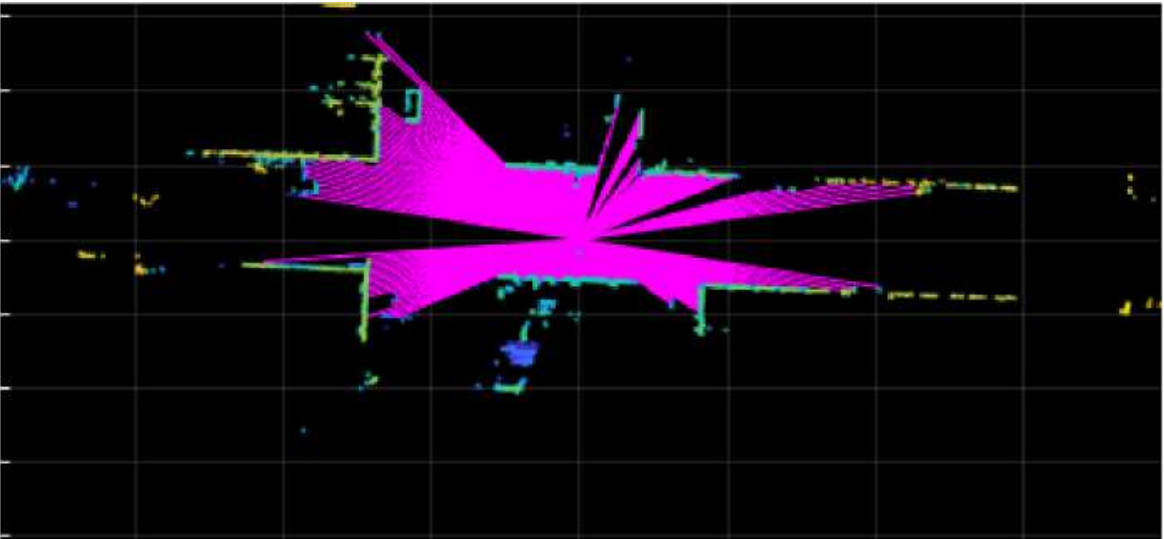}} \\
\subfigure[Top 1]{\includegraphics[width=0.18\linewidth]{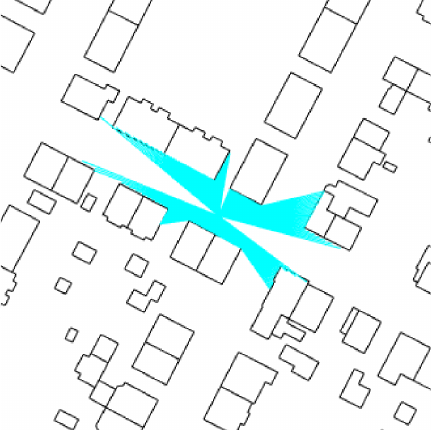}}
\subfigure[Top 2]{\includegraphics[width=0.18\linewidth]{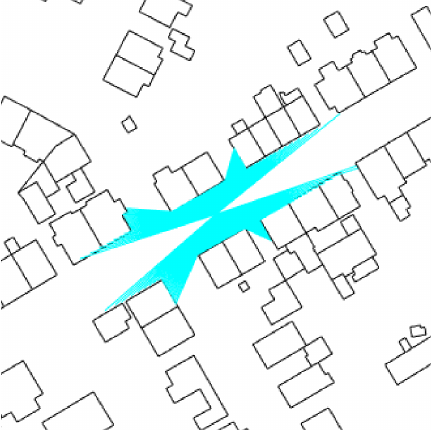}}
\subfigure[Top 3]{\includegraphics[width=0.18\linewidth]{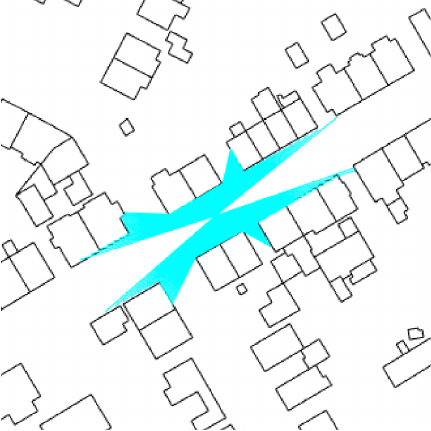}}
\subfigure[Top 4]{\includegraphics[width=0.18\linewidth]{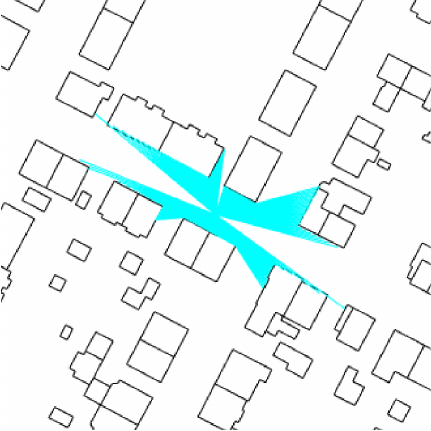}}
\subfigure[Top 5]{\includegraphics[width=0.18\linewidth]{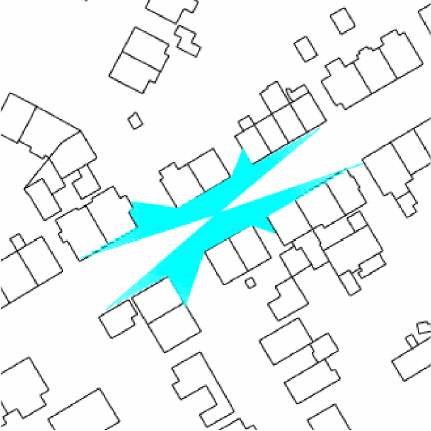}}
\caption{Top 5 reference OSM descriptors against query LiDAR point cloud.}
\label{fig:des_result}
\end{figure}

First, we evaluated the performances on twelve KITTI and KITTI-360 datasets using the top $N$ candidates. In this paper, we used $N = 200$. \figref{fig:kitti_results} and \figref{fig:kitti360_results} illustrate the qualitative results of the KITTI and KITTI-360 dataset sequences, respectively. \figref{fig:des_result} shows the top five candidates for the query LiDAR point cloud. It shows our descriptor's rotation-invariant characteristic, because they find the right reference OSM descriptors even when the query LiDAR descriptor and reference descriptor's directions are different.

The evaluation was performed using every query LiDAR scan in the KITTI and KITTI-360 dataset sequences, even if a scan revisits the same place multiple times. This suggests robustness against rotation, because our OSM keys are rotation-invariant. Especially, the result at KITTI-360 sequence 00 shows our method's rotation-invariant performance, indicating over 95\% accuracy at intersections, where the vehicle passed from at least two different directions. In practice, autonomous vehicles usually follow a lane in normal situations and make decisions at intersections, where they can only change their direction. Therefore, the ability to distinguish intersections is important for autonomous driving.

The quantitative results of our method are presented in \tabref{tab:KITTI_result}. The magnitude of the difference between the top one and top five is similar to that between the top five and top 10. This demonstrates that our method can yield as effective results even with a lower number of candidates. Several sequences show poor performance, which we will analyze in Section \ref{chap:failure_cases}.

\subsection{Comparison with PointNetVLAD}
\begin{figure}[!t]
\centerline{\includegraphics[width=0.8\linewidth]{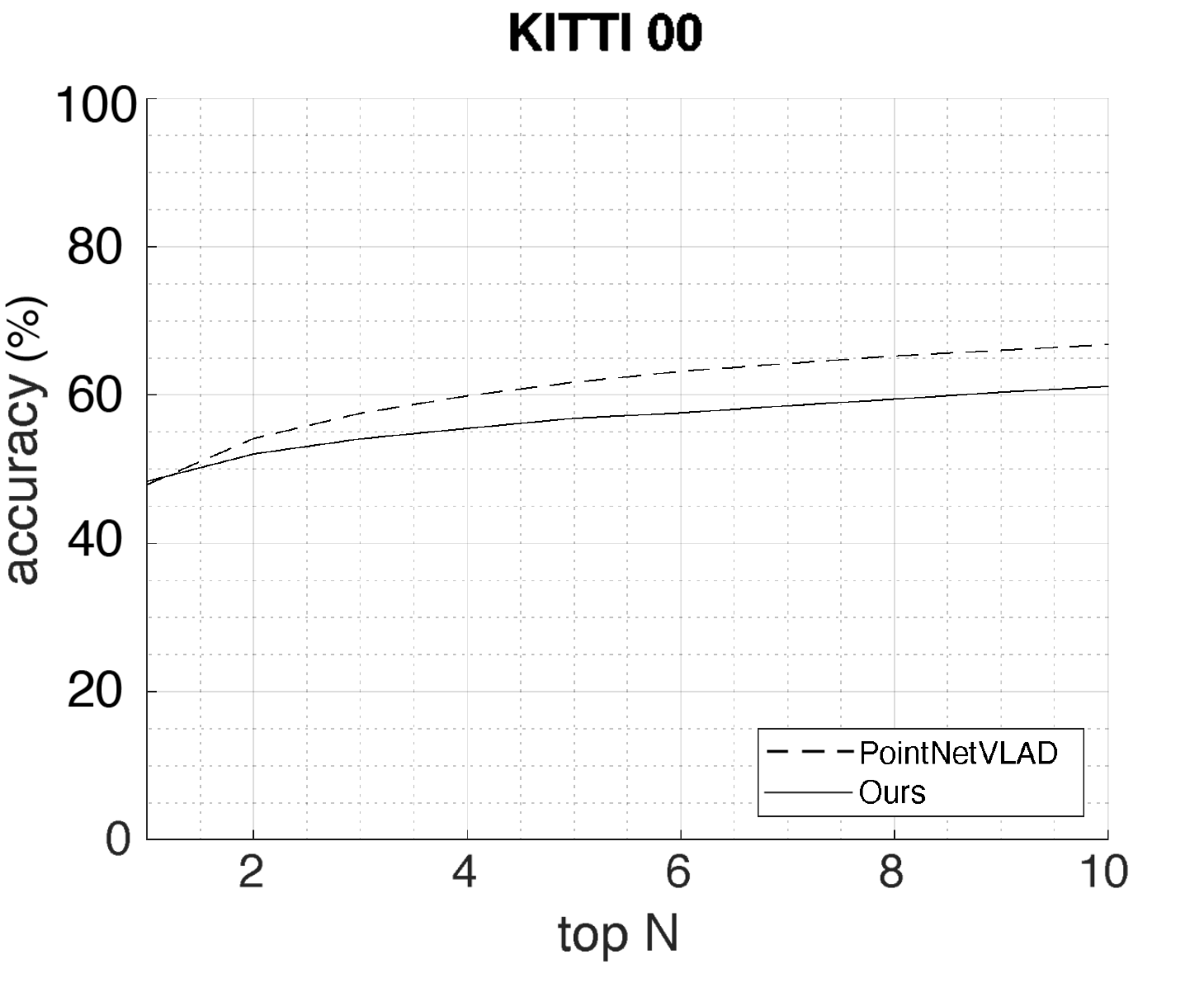}}
\caption{Top 10 accuracies for our method and PointNetVLAD. Our method performs comparable global localization without a prior LiDAR map and ourperforms with the top one candidate.}
\label{fig:kitti00_compare}
\end{figure}

\tabref{tab:pnv_compare} and \figref{fig:kitti00_compare} show our method's performance compared with PointNetVLAD. The KITTI dataset sequence 00 was used for comparison, as mentioned in Section \ref{chap:comparison_methods}. Our method shows a comparable result to PointNetVLAD, even though we did not utilize LiDAR maps and deep-learning networks. In particular, our method outperforms PointNetVLAD at the top one accuracy.

\subsection{Runtime Analysis}

The OSM descriptor is also a lightweight descriptor compared to the PointNetVLAD. Our method consists of two stages; therefore, we compare the performance and runtime for 1-stage and 2-stage results. \tabref{tab:calc_time} shows the runtimes for each method executed in the same hardware setting except for GPU usage in PointNetVLAD. All experiments were carried out on a PC with an Intel i7-6700 CPU at 3.40$\,$GHz and 32$\,$GB memory. For PointNetVLAD experiments, we used an additional GeForce GTX 1080 Ti GPU. The localization time is proportional to the size of the descriptor, because we used the same comparison algorithm for all the descriptors. The sizes of the descriptors are 1-by-256, 1-by-10, and 1-by-360 for PointNetVLAD, OSM keys, and OSM descriptors, respectively. However, the descriptor generation time is comparable with PointNetVLAD, even though PointNetVLAD was processed on a GPU.

\begin{figure}[!t]
\centerline{\includegraphics[width=0.8\linewidth]{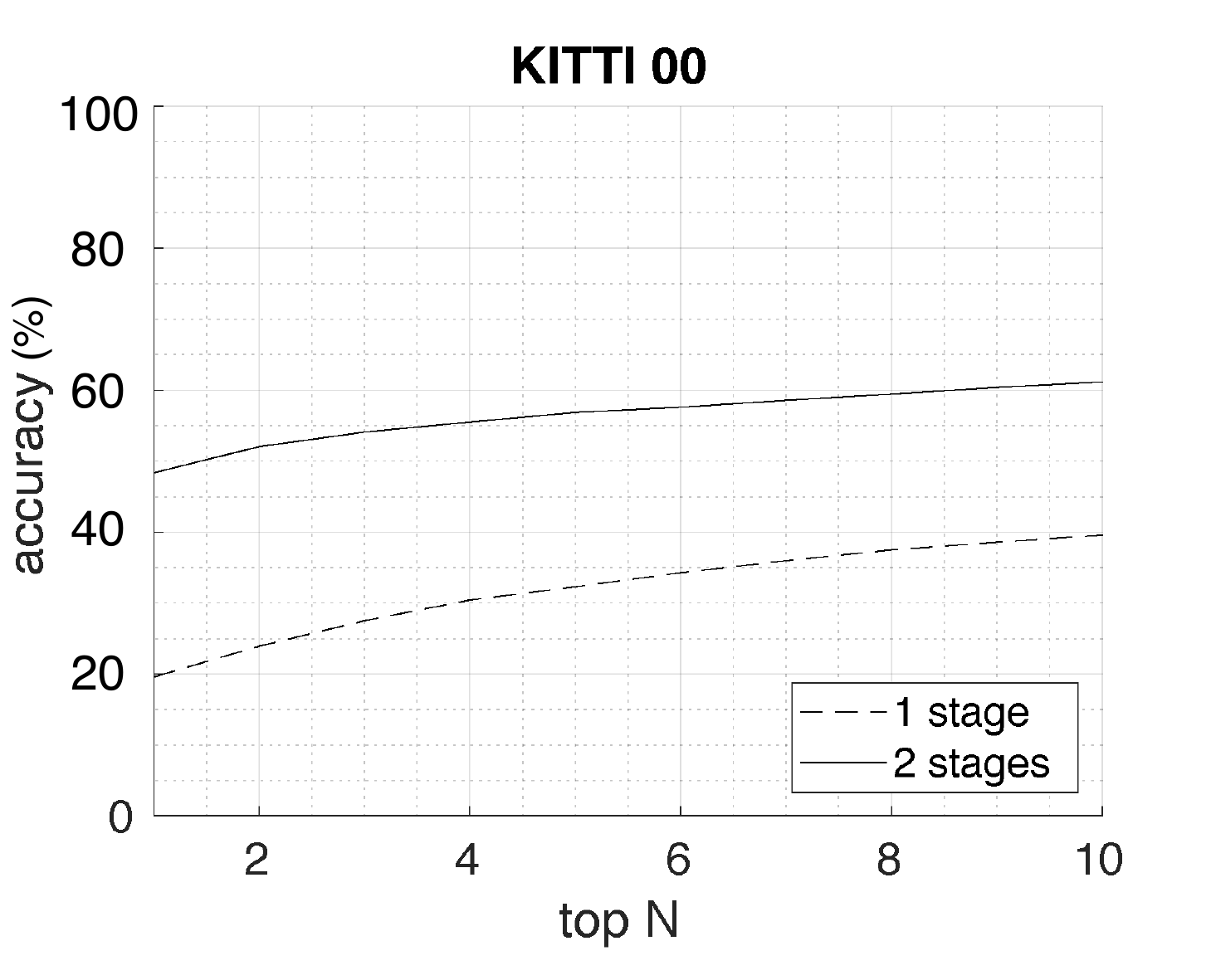}}
\caption{Global localization accuracy of the 1-stage method and 2-stage method.}
\label{fig:2stages_acc}
\end{figure}

Additionally, \tabref{tab:calc_time} shows that there is a 0.0534$\,$s difference between the 1-stage method and 2-stage method. This means that it took 0.0534$\,$s to compare the query descriptor and the top 200 OSM descriptors. If we compare all descriptors with out-of-sorting in stage 1, it might take $1.872 = 0.0534 * (7010 / 200)$ more seconds for every query LiDAR scan, which takes 8,499 more seconds for the entire KITTI dataset sequence 00. \tabref{tab:2_stage_acc} and \figref{fig:2stages_acc} compares the accuracy of the 1-stage method and 2-stage method.
\begin{table}[!t]
\caption{Performance against PointNetVLAD (\%)}
\begin{center}
\begin{tabular}{c|ccc}
\hline
method & Top 1 & Top 5 & Top 10 \\
\hline
\textbf{PointNetVLAD} &  47.90 & \textbf{61.73} & \textbf{66.79} \\
\hline
\textbf{Ours} & \textbf{48.34} & 56.86 & 61.15 \\
\hline
\end{tabular}
\label{tab:pnv_compare}
\end{center}
\end{table}

\subsection{Failure Cases}
\label{chap:failure_cases}

\begin{figure*}[!t]
\centerline{\includegraphics[width=0.95\textwidth]{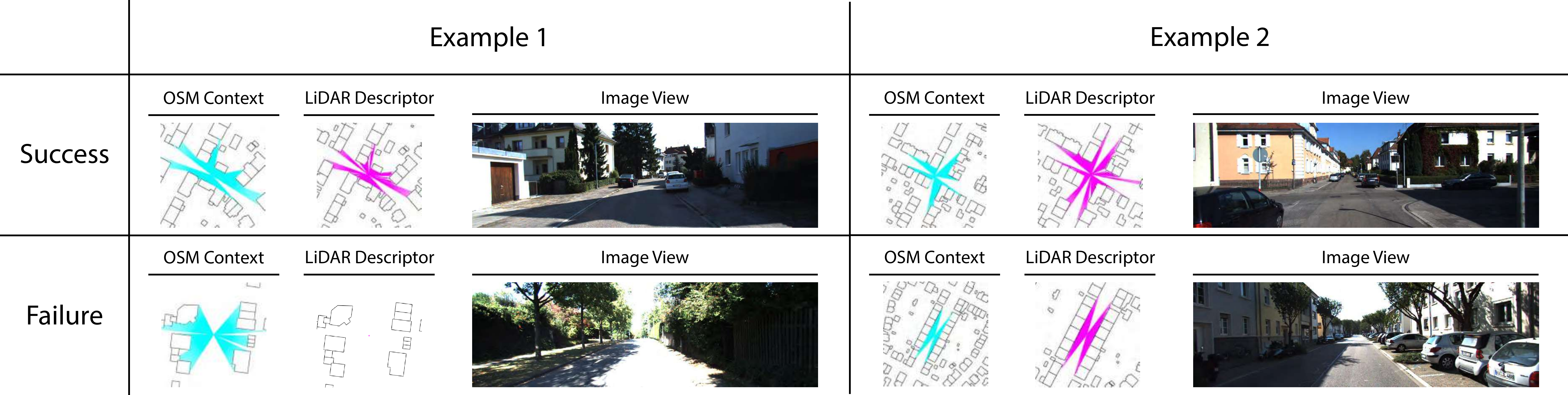}}
\vspace{-0.2cm}
\caption{Some sequences showed poor localization accuracy, with our method being weak in three situations: first, as shown in Example 1, if the buildings are blocked by fences, trees, or hills, there is no building information in LiDAR scans, which generates a blank descriptor; second, as shown in Example 2, if there are repeated building patterns, there are multiple candidates that have similar descriptors, which decreases the discrimination ability of our descriptors; finally, if there is no building around a vehicle, such as a highway or forest, a building-based descriptor cannot be made.}
\vspace{-0.4cm}
\label{fig:failure_case}
\end{figure*}

As mentioned above, several sequences showed poor performances. In this subsection, we analyze the failure cases in three scenarios. In the first scenario, as shown in failure example 1 in \figref{fig:failure_case}, the building is blocked by fences and trees or the building is on a hill. In this case, the vehicle perceives that there is no building around the vehicle, but the OSM descriptor is surrounded by buildings. Dynamic point removal algorithms such as ERASOR \cite{lim2021erasor} can help resolving this problem. In the second case, a similar building pattern was repeated over a long interval. In failure example 2, as shown in \figref{fig:failure_case}, both sides of the vehicle are full of buildings without a gap between the buildings. Therefore, our descriptor loses its discrimination power in these areas. In the last case, if there are no buildings around the vehicle, such as a highway or off-road, the building-based descriptor cannot be made, and our method does not work.

\begin{table}[!t]
\caption{Runtime comparison per frame}
\begin{center}
\begin{adjustbox}{width=1\columnwidth}
\begin{tabular}{c|ccc}
\hline
& \textbf{PointNetVLAD} & \textbf{Ours (1-Stage)} & \textbf{Ours (2-Stage)} \\
\hline
Descriptor Generation & \textbf{0.0690s} & 0.1714s & 0.1714s \\
Localization & 0.0167s & \textbf{0.0090s} & 0.0624s \\
\hline
Total & \textbf{0.0857s} & 0.1804s & 0.2338s \\
\hline
\end{tabular}
\end{adjustbox}
\label{tab:calc_time}
\end{center}
\end{table}

\begin{table}[!t]
\caption{Accuracy of 2-Stage Method (\%)}
\begin{center}
\begin{tabular}{c|ccc}
\hline
method & Top 1 & Top 5 & Top 10 \\
\hline
\textbf{1 stage} & 19.60 & 32.31 & 39.60 \\
\hline
\textbf{2 stages} & \textbf{48.34} & \textbf{56.86} & \textbf{61.15} \\
\hline
\end{tabular}
\label{tab:2_stage_acc}
\end{center}
\end{table}

\section{Conclusion}
\label{chap:conclusion}

In this study, we proposed a localization method using the OpenStreetMap information which can be executed without any prior LiDAR maps. We showed that our method performs comparably with PointNetVLAD, which uses LiDAR maps and deep learning techniques, in terms of accuracy and efficiency. The newly proposed novel descriptor, context, and key for OSM and LiDAR data enables a fast and highly accurate localization performance invariant to rotation. We verified its performance on 12 sequences of the KITTI dataset and KITTI-360 dataset. Our method outperforms PointNetVLAD in terms of the top one accuracy at KITTI dataset sequence 00, and showed comparable performance in top five and ten accuracy.
However, in KITTI sequence 02, buildings are blocked by other environments; therefore, our method does not perform well because a LiDAR scan measures walls around the vehicle, while the OSM descriptor recognizes that the same place is surrounded by buildings. In future work, we aim to improve our method to function in challenging environments with sparse buildings, such as the KITTI sequence 02 and MulRan sequence ‘Sejong city’ \cite{kim2020mulran}.

\addcontentsline{toc}{chapter}{Bibliography}
\renewcommand*{\bibfont}{\small}
\bibliographystyle{IEEEtranN} 
\bibliography{reference}

\end{document}